\documentclass[11pt]{article}

\usepackage[preprint]{acl}

\usepackage{times}
\usepackage{latexsym}
\usepackage{multirow}
\usepackage{subcaption}
\usepackage{adjustbox}
\usepackage{amsmath}
\usepackage{amssymb}
\usepackage{mathtools}
\usepackage{amsthm}
\usepackage{bbm}
\usepackage{xcolor}
\usepackage{booktabs}
\usepackage[breakable]{tcolorbox}
\newtcolorbox{promptbox}[1]{ 
  colback=gray!10,
  colframe=gray!50,
  title=#1,
  halign title=center,
  breakable,
  fontupper=\scriptsize
}

\usepackage[T1]{fontenc}

\usepackage[utf8]{inputenc}

\usepackage{microtype}

\usepackage{inconsolata}

\usepackage{graphicx}

\title{ProPlay: Procedural World Models for Self-Evolving LLM Agents}

\author{
    \textbf{Yijun Ma}$^{1,*}$\;
    \textbf{Zehong Wang}$^{1,*,\dagger}$\;
    \textbf{Yiyang Li}$^{1}$\;
    \textbf{Ziming Li}$^{2}$\; \\
    \textbf{Xiaoguang Guo}$^{2}$\; 
    \textbf{Weixiang Sun}$^{1}$\;
    \textbf{Chuxu Zhang}$^{2}$\;
    \textbf{Yanfang Ye}$^{1,\dagger}$
    \\
    \textsuperscript{1} University of Notre Dame\; 
    \textsuperscript{2} University of Connecticut\; 
    \\
    $^*$ Equal Contribution\;
    $^\dagger$ Corresponding Author
    \\
    \texttt{<yma7,zwang43,yye7>@nd.edu}
}

\begin{document}
\maketitle

\begin{abstract}


Self-evolving agents are expected to improve through interaction without external supervision, but this remains difficult in partially observable environments where agents must explore actively, learn from limited feedback, and decide when to trust prior experience. Existing LLM-agent methods often rely on memory or planning modules, yet they rarely close the loop between them to continually refine an internal understanding of environment dynamics. We introduce ProPlay, a procedural world model that supports procedure-level preplay, where agents can rehearse future procedural paths using the learned world knowledge. Rather than representing experience as isolated rules or low-level action constraints, ProPlay abstracts successful trajectories into procedures and organizes them in a procedure graph that captures causal transitions among task stages. Each transition is associated with a reliability record embedding to estimate its task-specific contribution from past outcomes. Before each episode, ProPlay simulates future procedural trajectories over known graph structures as structured soft guidance; after execution, it refines the graph using environment feedback. Experiments on public benchmarks show that ProPlay consistently improves environment understanding and self-evolution capability over strong baselines. Our code has been released in \url{https://github.com/antman9914/proplay}.

\end{abstract}

\section{Introduction}

\begin{figure}[!t]
    \centering
    \includegraphics[width=\linewidth]{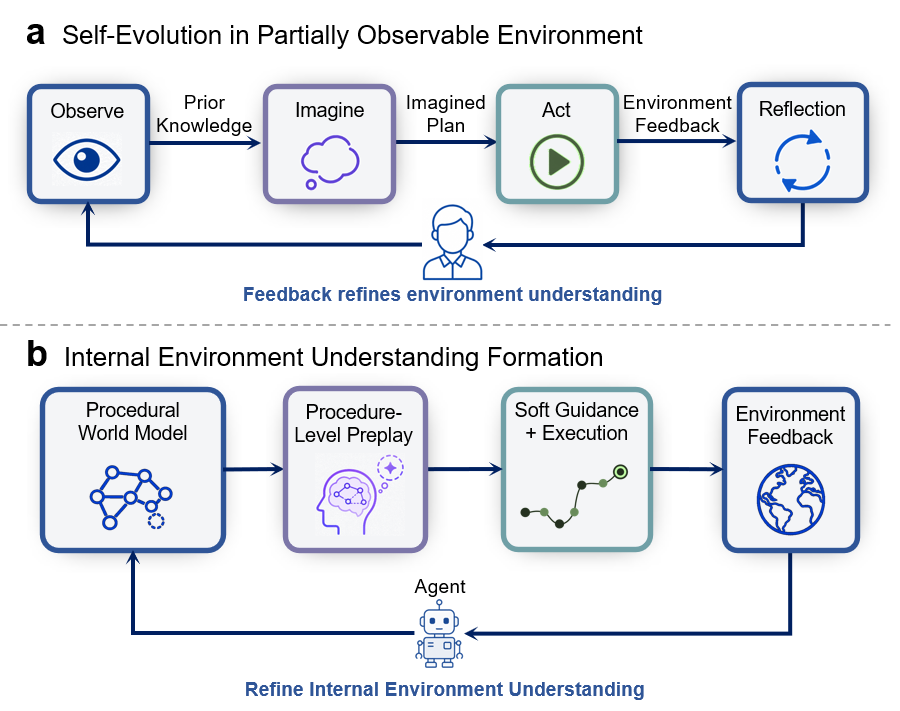}
    \caption{\textbf{Motivation of ProPlay.} (1) Humans adapt to partially observable environments by imagining plausible future paths before acting and consolidating or discarding them after receiving feedback. (2) We expect self-evolving agents to follow a similar loop: use prior experience to anticipate procedural paths, test them through interaction, and refine internal environment understanding with environment feedback.}
    \label{fig:motivation}
\end{figure}

An important goal of Agentic AI is to build agents that improve through their own interactions, without relying on additional supervised data or large-scale retraining~\cite{fang2025comprehensive}. Such self-evolving agents should be able to acquire new capabilities, refine their strategies, and adapt to unfamiliar environments from experience~\cite{gao2025survey,wang2023voyager}. This goal is especially challenging in partially observable environments~\cite{kaelbling1998planning}, where an agent rarely observes the full state of the world and must infer environment dynamics from sparse, delayed, and sometimes misleading feedback.

A useful way to view this challenge is through the interaction between imagination and consolidation. Humans often anticipate possible future paths before acting, and later use feedback to reinforce or discard those expectations. Neuroscience studies describe a related mechanism known as preplay, where hippocampal activity can pre-enact candidate future trajectories before execution~\cite{dragoi2011preplay,olafsdottir2018role}. This process naturally connects two capacities that are central to self-evolution: planning, which uses current knowledge to imagine what to do next, and memory, which consolidates useful experience for future decisions~\cite{schacter2007remembering}. For LLM agents, the key question is therefore not only how to store past experience or generate plans, but how to close the loop so that each interaction improves the agent's internal understanding of the environment~\cite{sajid2021active,li2026graph}.

Closing this loop requires solving three coupled problems. (1) The agent must explore actively, since passive observation is insufficient for recovering hidden environment dynamics~\cite{pezzulo2024generating}. (2) It must accumulate experience at the right level of abstraction: real tasks often involve hierarchical, long-horizon, and causally structured progressions~\cite{yu2023explainable}, so raw trajectories or isolated rules may not reveal why a task succeeds. (3) It must decide when to trust previous experience. Past behavior can be useful, but it can also be noisy, overfitted to particular task instances, or irrelevant to the current goal. Effective self-evolution thus requires a mechanism that can acquire, refine, and selectively utilize environment knowledge through repeated interaction~\cite{friston2017active}.

Existing LLM-agent methods address parts of this problem, but often leave the loop incomplete. Memory-based approaches store reflections, rules, or workflows from previous episodes, while planning-based approaches search over possible future actions or reasoning paths~\cite{zhao2024expel,zhou2024language,wang2025agent,wang2026reasoning,li2026same,zhang2026semantic,wu2026proteo,zhang2025agentrouter}. These components are useful, yet they are typically treated as separate modules rather than as a unified model of how the environment evolves. World models offer a natural alternative, because they explicitly represent how actions, states, and outcomes are related~\cite{ha2018world,hafner2019dream,hafner2023mastering}. Recent LLM-agent systems such as Wall-E~\cite{zhou2026wall} and WorldCoder~\cite{tang2024worldcoder} take this direction by learning action-level rules or executable functions from interaction history. However, action-level world models mainly capture local constraints. They can miss the higher-level causal structure of a task, and their direct influence on action selection can also make it difficult to balance exploitation with continued exploration.

We argue that self-evolving agents need world models that capture \textit{procedural causal structure}. A procedure is a high-level abstraction of a task stage, induced from successful interaction trajectories, that describes what role the stage plays and how it can be carried out. Compared with individual actions or local rules, procedures represent reusable progress patterns such as boiling water, measuring a property, assembling required tools, or verifying a result. Transitions among procedures then describe how task stages depend on one another. This procedural view gives the agent a structured prior over long-horizon task progress while preserving flexibility in low-level action choices~\cite{wang2023describe}.

Based on this perspective, we introduce \textbf{ProPlay}, a preplay framework built on an evolving procedural world model for LLM agents. ProPlay represents environment knowledge as a procedure graph, where nodes are induced procedures and directed edges encode observed procedural transitions. To support reliable use of past experience, each transition maintains a reliability record embedding that reflects how consistently the transition contributed to successful tasks with similar descriptions. Before each new episode, ProPlay performs procedure-level preplay: it reasons over the current task, the procedure graph, transition reliabilities, and relevant failure experience to construct a task-specific procedural trajectory. This trajectory is injected into the agent context as soft guidance rather than a hard constraint, allowing the agent to exploit accumulated knowledge while still adapting to observations and exploring unseen transitions. After execution, ProPlay refines the graph using environment feedback by inducing new procedures, updating transitions, and revising reliability records.

We evaluate ProPlay on three public interactive benchmarks: ScienceWorld, $\tau$-Bench, and PlanCraft. 
The experimental results show that procedure-level abstraction provides a useful foundation for agents that must learn environment dynamics online, especially in tasks with reusable multi-step causal structure.
Our main contributions are summarized:
\begin{itemize}
    \item \textbf{Procedure-Centric Self-Evolving Agent.} We propose a procedure-centric perspective for self-evolving LLM agents, which models environment dynamics as procedural pathways rather than low-level action constraints.
    \item \textbf{ProPlay.} We introduce a preplay framework that constructs and continually maintains a procedural world model from interaction experience, retrieving task-relevant procedural trajectories as soft guidance for LLM agents.
    \item \textbf{Empirical Evaluation.} We conduct experiments across three public benchmarks and demonstrate that ProPlay consistently improves environment understanding and self-evolution capability against strong baselines.
\end{itemize}

\section{Related Work}
\noindent

\paragraph{Self-Evolving Agent.}
Self-evolving agents autonomously improve through environment interaction without external supervision~\cite{gao2025survey,fang2025comprehensive}. A prominent thread focuses on memory-based experience accumulation, progressing from verbal reflection and rule extraction~\cite{shinn2023reflexion,zhao2024expel,wang2023voyager} to  persistent and hierarchical memory management~\cite{packer2023memgpt,kang2025memory,xu2026mem,wang2025agent}. Deliberate planning has advanced in parallel, from multi-path reasoning~\cite{yao2023tree} to tree-search-based decision making~\cite{zhou2024language} and world-model-guided exploration~\cite{hao2023reasoning,putta2024agent}. Despite these advances, existing approaches largely treat memory and planning as independent modules, failing to continually refine environment understanding under a unified framework.

\paragraph{World Model for LLM Agent.}
World models provide a principled framework for capturing environment dynamics to support forward-looking decision making. Early applications in LLM agents treat world knowledge as a static external prior — leveraging pre-trained commonsense knowledge~\cite{zhao2023large} or curated action knowledge bases~\cite{zhu2025knowagent} to guide planning. RAP~\cite{hao2023reasoning} marked a conceptual turning point by casting the LLM itself as an implicit world model within MCTS, demonstrating the value of internal forward simulation for planning quality. This inspired a growing line of work building internal world models from agent experience: WKM~\cite{qiao2024agent} distills parametric world knowledge from offline trajectories, while WorldCoder~\cite{tang2024worldcoder} and Wall-E~\cite{zhou2026wall} enable online updates via code-based and rule-based models, with recent extensions to web navigation~\cite{chae2025web}. However, all these methods overlook the procedural causal structure that governs holistic environment dynamics and long-range causal dependencies.
\begin{figure*}[!t]
    \centering
    \includegraphics[width=\linewidth]{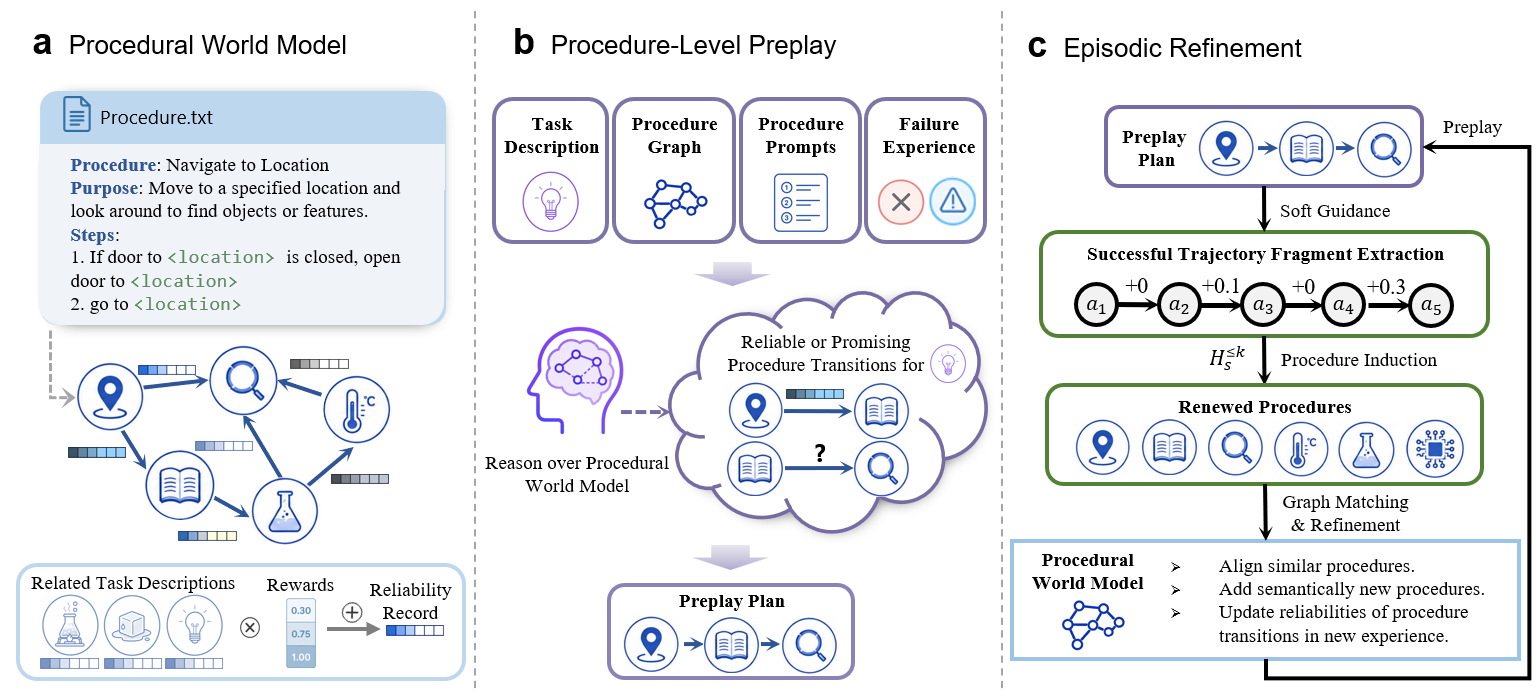}
    \caption{\textbf{Overview of ProPlay.} (1) Procedural world model represents environment dynamics via a procedure graph with reliability records assigned for each procedure transition. (2) Procedure-level preplay constructs procedural plan by reasoning over existing world knowledge and current task description. (3) Preplay plan functions as soft guidance for reasoning, and the resulting action trajectory is collected for further procedure induction and world model refinement, which forms an episodic execution loop.}
    \label{fig:method}
\end{figure*}

\section{Methodology}

\paragraph{Problem Formulation.}
Given a partially observable environment with observation space $\mathcal{O}$ and action space $\mathcal{A}$, a self-evolving agent is expected to continually adapt to the environment across a sequence of $K$ task episodes $\{\tau_i\}_{i=1}^K$. At each step $t$ within episode $k$, the agent receives an observation $o_t \in \mathcal{O}$, selects an action $a_t \in \mathcal{A}$, and receives reward $r_t \in \mathbb{R}$. The trajectory of episode $k$ is denoted $H^k = {(o_1, a_1, r_1), \ldots, (o_T, a_T, r_T)}$, and $R^k \in [0, 1]$ indicates the final reward episode $k$ obtained. Each task episode instance $\tau$ is associated with a natural language description $d_\tau$. The agent is expected to maximize task success rate and total reward $\frac{1}{K}\sum_{k=1}^K R^k$ through continual interactions with environment.

\paragraph{Overview.}
As illustrated in Figure \ref{fig:method}, the core of ProPlay is a continually evolving procedure graph $\mathcal{G}$ that represents environment dynamics as a collection of abstracted procedures, their causal transitions and corresponding reliability record embeddings, which is essentially a procedural world model. Proplay operates in an episode-level three-phase loop. Before each episode, a preplay mechanism generates the most relevant procedure trajectory from $\mathcal{G}$ and provides it as structured soft guidance to the LLM agent. During the episode, the agent executes under this guidance, maintaining the flexibility to deviate and explore. After completion, environment feedback is collected to refine $\mathcal{G}$: new procedures are induced from successful executions, and procedure transition reliabilities are adjusted based on received reward. This closed loop enables ProPlay to progressively internalize environment dynamics over episodes.

\subsection{Procedural World Model}

Internal environment understanding is the foundation of both environment exploration and experience exploitation. Building such a representation is non-trivial, as environment dynamics are partially observable, hierarchically structured, and causally complex~\cite{kaelbling1998planning}. Prior approaches tend to adopt rules~\cite{shinn2023reflexion}, individual actions~\cite{zhou2026wall,tang2024worldcoder} and entities~\cite{anokhin2025arigraph} as basic unit of environment representation, but they fail to express the holistic stage-wise causal structure underlying task progression.

Instead, ProPlay represents environment dynamics using procedures as the fundamental unit of the world model. Rather than specifying individual actions, a procedure $p$ is natural language description of the purpose and execution template of a generalizable task stage, which is abstracted from successful interaction trajectories. Procedure-level abstraction allows the world model to represent the skeleton of task progression in a form that generalizes across task instances while remaining grounded in actual agent experience.

These procedures are organized into an attributed directed graph $\mathcal{G} = (\mathcal{P}, \mathcal{E}, \mathcal{C})$, where $\mathcal{P}$ is the set of procedure nodes, and $\mathcal{E} \subseteq \mathcal{P} \times \mathcal{P}$ is the set of directed edges encoding causal transitions among task stages. A key challenge in accumulating experience from partially observable environments is that interactions may reflect contingent or suboptimal behavior rather than actual environmental regularities. Therefore, we introduce reliability record embedding $\mathcal{C}: \mathcal{E} \rightarrow \mathbb{R}^{d_c}$ to each transition, where $d_c$ is the dimension of record embedding. The reliability record embedding is designed to quantify how consistently the given procedure transition has contributed to the success of specific tasks, in order to elevate generalizable procedural knowledge while suppressing misleading experience. Formally, given procedural transition $(p_i, p_j)$, its reliability record embedding $c_{ij}^k$ at the end of episode $k$ is defined as the weighted summation of relevant seen task description embeddings:

\begin{equation}
\label{eq:reliab}
    c_{ij}^k=\sum_{i=1}^k \mathbb{I}((p_i,p_j)\in W^i) \cdot R^i\cdot\phi(d_{\tau_i})
\end{equation}

where $W^i = (p_{i_1}, \ldots, p_{i_n})$ represents the executed procedure pathway at episode $i$, $\mathbb{I}(\cdot)$ is an indicator function indicating whether $(p_i,p_j)$ is part of $W^i$, $\phi(\cdot)$ is the \texttt{all-MiniLM-L6-v2} text encoder for task descriptions.

\subsection{Procedure-Level Preplay}

Effective self-evolution requires agents to be active, predicting about how a task should unfold and learning from its own errors~\cite{pezzulo2024generating}. This directly motivates our preplay mechanism: at the start of episode $k$, ProPlay queries procedure graph $\mathcal{G}^{k-1}$ derived from episode $k-1$ to construct a procedure trajectory over the current task $\tau_k$, which the agent then executes to confirm, refine, or revise through interaction.

\paragraph{Procedure Trajectory Construction.} ProPlay delegates trajectory construction to the agent itself. To learn from relevant failure and success, evidence provided to agent is composed of three parts: task description $d_{\tau_k}$, the full procedure graph $\mathcal{G}^{k-1}$, and failure experiences from past episodes. We first extract failed episodes with semantically close task description via cosine similarity between $\phi(d_{\tau_k})$ and $\phi(d_{\tau_i})$, and then assign each observed procedure transition $c_{ij}^{k-1}\in\mathcal{C}^{k-1}$ with a task-specific reliability score $s_{ij}^{k-1}$, which is defined as the cosine similarity between $\phi(d_{\tau_k})$ and $c_{ij}^{k-1}$.
All the observed procedure transitions in $\mathcal{E}^{k-1}$ are injected in the descending order of $s_{ij}^{k-1}$, so that task-relevant transitions surface prominently. 
LLM agent reasons over $d_{\tau_k}$ and all the injected evidence, constructing procedure trajectory $W^* = (p_1, \ldots, p_m)$ as the preplay plan for the current task. Note that unknown procedure transitions are allowed in $W^*$ to facilitate exploration.

\subsection{Episodic World Model Refinement}

During episode $k$, $W^*$ functions as the soft guidance of LLM agents, allowing agents to reason over low-level actions independently. The resulting trajectory $H^k$ constitutes the environment feedback against which the agent's preplay expectations can be compared. This expectation-feedback comparison is the mechanism by which ProPlay continuously refines its understanding of environment dynamics: where expectations align with outcomes, the world model is reinforced; where they diverge, the world model is updated to better reflect real environment dynamics.

\paragraph{Soft Guidance.} During each episode, $W^*$ is always injected into the agent's context prompt as a high-level structured prior. It is noteworthy that $W^*$ is provided as soft guidance rather than a hard constraint: the agent is explicitly permitted to deviate from the suggested pathway when observations warrant it. By preserving the agent's freedom to explore while grounding its reasoning in accumulated procedural knowledge, our preplay mechanism enables a principled balance between exploitation and exploration. Formally, the agent policy at step $t$ is conditioned on both the interaction context and the preplay trajectory, i.e., $a_t \sim \pi_\theta\left(a_t \mid o_{\leq t}, W^*\right)$.

\paragraph{Procedure Induction.} After episode $k$ terminates, ProPlay first partitions $H^k$ at the last step where cumulative reward increased. The prefix productive portion $H^k_s$ that contributed to task progress is passed to LLM agent for procedure induction, while the suffix portion $H^k_f$ that failed to yield further progress is summarized and stored as a failure experience associated with $d_{\tau_k}$. ProPlay extracts a new workflow set from all the past successful trajectories $H^{\le k}_s$ by prompting LLM agent to segment the trajectories into coherent task stages and induce a new set of procedures $\mathcal{P}_\text{new}$. To maintain a coherent and non-redundant procedural vocabulary, each $p_\text{new}\in\mathcal{P}_\text{new}$ is compared against existing procedures: if the semantic similarity between $p_\text{new}$ and any $p\in\mathcal{P}$ exceeds a threshold $\delta$, $p_\text{new}$ is merged with its nearest neighbor, integrating new experience and correcting potential errors; Otherwise, it is added as a new procedure, expanding the world model's knowledge of environment dynamics. After induction, $H^k_s$ is mapped to the updated $\mathcal{P}$, and the corresponding procedure pathway $W^k$ is derived, where new consecutive procedure transitions are added to $\mathcal{E}$.

Although our procedure induction method is similar to AWM~\cite{wang2025agent}, they serve fundamentally different goals. AWM constructs a library of independently reusable workflow templates as episodic memory, while ProPlay treats induced procedures as basic units in an evolving world model, where the procedures and their causal transition structures constitute the accumulated understanding of the environment. 

\paragraph{Reliability Record Update.} The reliability record embedding of each procedure transition observed in episode $k$ is updated based on Equation \ref{eq:reliab} to reflect their task-specific contributions. 
This ensures that reliability scores can consistently track the task-specific empirical reward gain of each transition, consolidating generalizable knowledge and gradually discarding noisy transitions. As a result, procedural world model can provide more grounded signal for further exploration.

The episodic query-execute-refine loop ensures that ProPlay always benefits from up-to-date experience, while maintaining a balance between exploitation and exploration.

\begin{table*}[!t]
  \centering
  \resizebox{2.0\columnwidth}{!}{
    \begin{tabular}{c|cc|ccc|cccc|c}
    \toprule
    \multirow{2}[3]{*}{\textbf{Methods}} & \multicolumn{2}{c|}{\textbf{ ScienceWorld }} & \multicolumn{3}{c|}{\textbf{$\tau$-Bench}} & \multicolumn{4}{c|}{\textbf{PlanCraft}} & \multirow{2}[3]{*}{\textbf{A.R.}} \\
\cmidrule{2-10}          & \textbf{ SR } & \textbf{Avg. Score} & \textbf{Retail} & \textbf{Airline} & \textbf{Overall} & \textbf{Easy} & \textbf{Medium} & \textbf{Hard} & \textbf{Overall} &  \\
    \midrule
    ReAct & 27.0  & 58.7  & 80.9  & 84.0  & 81.8  & 61.3  & 45.0  & \underline{7.5}   & 38.5  & 5.3 \\
    Reflexion & 32.2  & 65.0  & \underline{81.7}  & \textbf{90.0} & 84.2  & 66.2  & \textbf{50.0}  & 9.0   & 42.2  &\underline{3.1} \\
    ExpeL & \underline{36.3}  & \underline{67.9}  & 80.0  & \textbf{90.0} & 83.0  & \textbf{72.5} & \textbf{50.0}  & 4.5   & \underline{43.3}  & 3.3 \\
    LATS  & 18.9  & 48.1  & \underline{81.7}  & 82.0  & 81.8  & 70.0  & 45.0  & \textbf{10.4}  & \underline{43.3}  & 4.8 \\
    Wall-E & 24.8  & 52.3  & \textbf{83.5} & \underline{88.0}  & \underline{84.9}  &  63.7  &  45.0 & \underline{7.5} & 39.6 & 4.4 \\
    WorldCoder & 24.8  & 53.4  & \textbf{83.5} & 82.0  & 83.0  & 61.3  & 42.5 & \underline{7.5} & 38.0 & 5.4 \\
    \midrule
    ProPlay & \textbf{37.4} & \textbf{70.2} & \textbf{83.5} & \textbf{90.0} & \textbf{85.5} & \underline{71.2}  & \underline{47.5}  & \textbf{10.4}  & \textbf{44.4}  & \textbf{1.6} \\
    \bottomrule
    \end{tabular}%
    }
    \caption{\textbf{Performance and Average Ranking (A.R.) Comparison.} Results are reported on ScienceWorld, $\tau$-Bench and PlanCraft, with the best and sub-best performance highlighted in \textbf{Boldface} and \underline{Underline}.}
  \label{tab:main}%
\end{table*}%

\begin{figure*}[!t]
    \centering
    \includegraphics[width=0.9\linewidth]{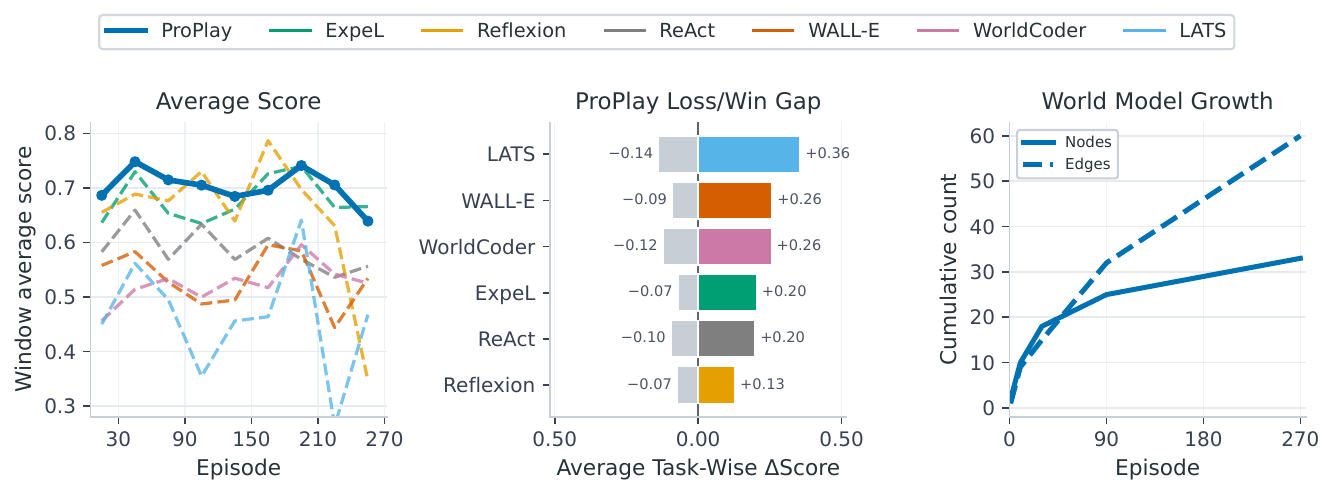}
    \caption{\textbf{Evolution Trend Analysis in ScienceWorld.} (1) ProPlay exhibits stable reward gain and advantages in cold-start stage. (2) ProPlay demonstrates significant advantages in tasks where it excels, while not falling noticeably behind in average when it comes to the tasks ProPlay is less proficient. (3) The growth in the scale of procedural world model is gradually stagnating, while the number of procedural transitions continues to increase. }
    \label{fig:evolution}
\end{figure*}

\section{Experiments}

\subsection{Experimental Setting}
\noindent

\paragraph{Benchmark.} We evaluate our ProPlay on three public benchmarks. ScienceWorld~\cite{wang2022scienceworld} is a text-based scientific reasoning environment. We evaluate over 270 shuffled tasks covering 23 task types under standard difficulty. $\tau$-Bench~\cite{yao2024tau} is a conversational benchmark that simulates realistic multi-turn tool-use interactions with an LLM-simulated user. We evaluate on both retail and airline domains, with 115 and 50 tasks respectively. PlanCraft~\cite{dagan2024plancraft} is a text-based crafting benchmark in a Minecraft-like environment that requires multi-step recipe planning and inventory management. We evaluate on 187 tasks stratified across three difficulty levels (easy/medium/hard). These benchmarks are used to evaluate different capabilities of ProPlay: $\tau$-Bench for single task adaptation, ScienceWorld for evolutionary ability in a complex multi-task environment, and PlanCraft for self-evolution under curriculum setting. 

\paragraph{Baseline.} We compare ProPlay with standard agents (ReAct~\cite{yao2022react}, Reflexion~\cite{shinn2023reflexion}), memory- and planning-augmented agents (ExpeL~\cite{zhao2024expel}, LATS~\cite{zhou2024language}) and world model-based agents (WorldCoder~\cite{tang2024worldcoder}, Wall-E~\cite{zhou2026wall}). ReAct and Reflexion evaluate whether reasoning and experience summarization can support self-evolution. ExpeL and LATS examine whether memory and planning can replace environment understanding. Wall-E and WorldCoder verify whether action-level world modeling is sufficient for self-evolution. All methods are evaluated under online inference setting, where no offline training is involved. Each task has only one trial. 
GPT-4.1-mini with temperature 0 is adopted as the backbone LLM. 

\paragraph{Metric.} For ScienceWorld, we report success rate (SR) and average reward (Avg. Score), where SR is the fraction of episodes completed successfully. For $\tau$-Bench, we report per-domain SR and overall SR. For PlanCraft, we report per difficulty level SR and overall SR.

\begin{figure*}[!t]
\centering
\begin{minipage}[!t]{0.57\textwidth}
\includegraphics[width=\linewidth]{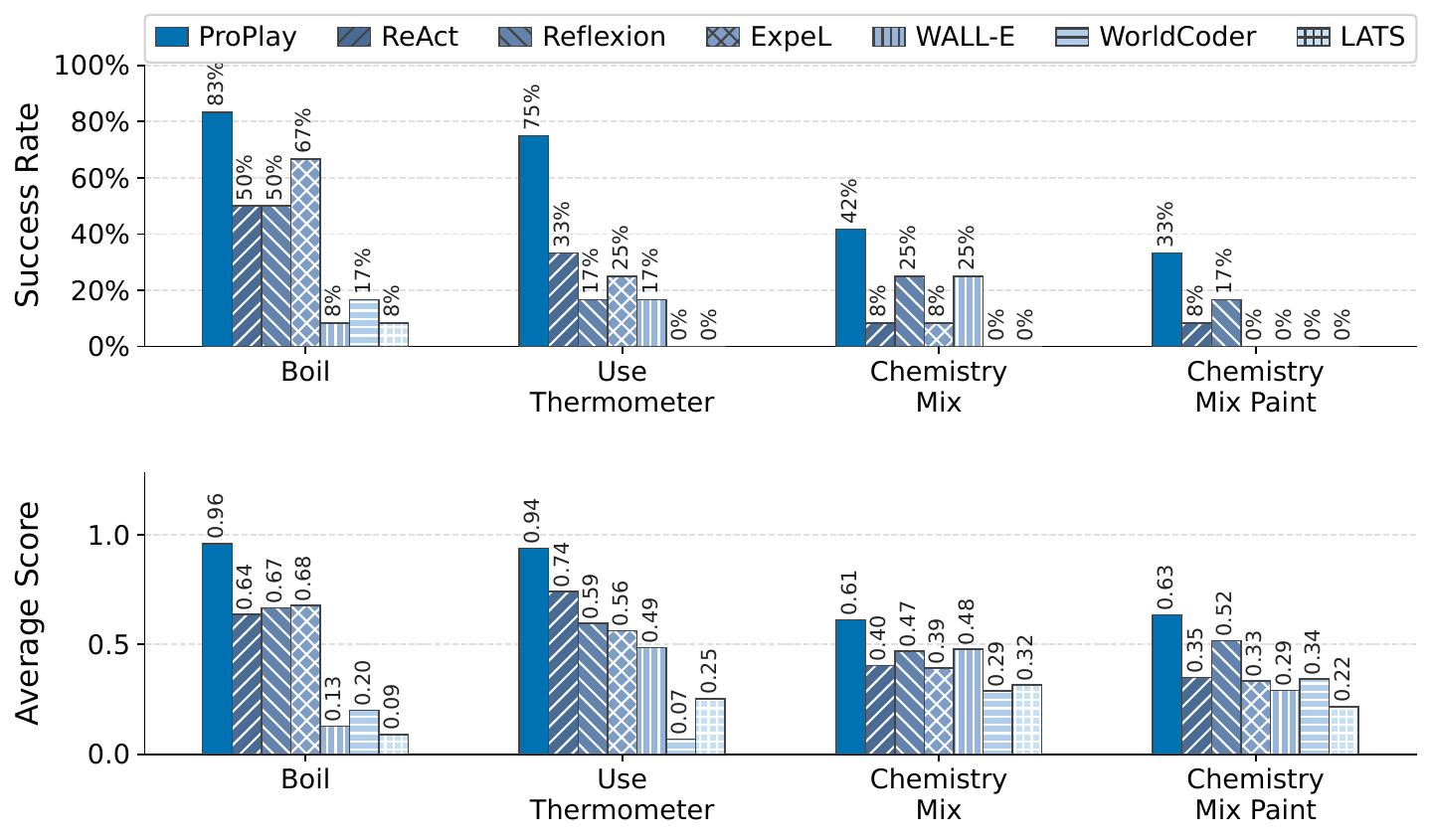}
\caption{\textbf{Task-Specific Results in ScienceWorld.} ProPlay demonstrates consistent advantages on tasks with clear multi-step procedural structure.}
\label{fig:task_bar}
\end{minipage}
\hfill
\begin{minipage}[!t]{0.42\textwidth}
\centering

  \resizebox{0.92\columnwidth}{!}{
    \begin{tabular}{l|cc}
    \toprule
    \multicolumn{1}{c|}{\multirow{2}[4]{*}{\textbf{Methods}}} & \multicolumn{2}{c}{\textbf{ScienceWorld}} \\
\cmidrule{2-3}          & \textbf{SR} & \textbf{Avg. Score} \\
    \midrule
    ProPlay & \textbf{37.4} & \textbf{70.2} \\
    \midrule
    w/ retrieval & 35.6  & 69.6 \\
    w/ random sample & 35.9  & 69.1 \\
    w/ hard constraint & 32.6  & 66.3 \\
    \midrule
    w/o graph & 34.8  & 69.4 \\
    w/o transition & 32.2  & 64.8 \\
    w/o reliability & 35.6  & 70.1 \\
    w/ action-level & 37.0  & 68.8 \\
    \bottomrule
    \end{tabular}%
    }
\captionof{table}{\textbf{Ablation of model components and alternative solutions in ProPlay on ScienceWorld.}
}
\label{tab:ablation}
\end{minipage}
\vspace{-10pt}
\end{figure*}

    


\subsection{Results}
\noindent

\paragraph{Main Results.}
Table \ref{tab:main} summarizes the online inference results. Overall, ProPlay achieves the best A.R., with best and sub-best performance over all the benchmarks. Notably, ProPlay significantly outperforms planning-based and world model-based baselines in ScienceWorld, demonstrating the superiority of our episodic preplay mechanism and procedure-level environment modeling in complex environments. ProPlay also achieves the best hard task SR after curriculum learning in PlanCraft, indicating that the procedural world model actually induced meaningful procedures from simpler tasks. Two world model-based baselines even fail to consistently outperform simple ReAct, indicating that their action-level constraints may significantly limit the reasoning flexibility of LLM agents. 

\paragraph{Self-Evolution Analysis.} 
Figure \ref{fig:evolution} presents the evolution trend of all the selected methods. (1) As shown in the evolution trend of average score over episodes, ProPlay demonstrates significant advantages in the early stages (before episode 90) and stable high reward gain across all the episodes. We attribute the observations to our preplay mechanism and soft procedural guidance design. During cold-start stage, the internal world model has not formed environment understanding, requiring agent to leverage its own prior knowledge to explore potential procedures. Later, the agent could also be limited by the world knowledge itself. Both periods require dynamic balance between exploration and exploitation, which other methods fail to achieve. (2) We also collect task-specific average score and compute the task-specific gap between ProPlay and other baselines. ProPlay demonstrates significant advantages in tasks where it excels, while not falling noticeably behind in average for less proficient tasks. It indicates that the accumulated procedural world knowledge can be generalized across heterogeneous tasks. (3) ProPlay's procedural graph exhibits a two-phase growth pattern analogous to cortical synaptogenesis~\cite{huttenlocher1979synaptic}: concept nodes accumulate rapidly in early episodes, while relational edges overtake and increasingly outnumber them afterwards. This dissociation suggests that deepening relational structure drives late-stage performance gains, which mirrors how synaptic density underlies the computational capacity of mature neural circuits~\cite{bullmore2009complex}.

\paragraph{Task-Specific Results.} Figure \ref{fig:task_bar} visualizes part of task-specific results on ScienceWorld, covering boil, use thermometer, chemistry mix and chemistry mix paint secondary color. ProPlay demonstrates strong and consistent advantages on tasks that require multi-step procedural reasoning with clear causal structure. On harder compositional tasks like chemistry mix and chemistry mix paint, ProPlay also generalizes across task variations and outperforms other baselines.

\subsection{Ablation Studies}

Table \ref{tab:ablation} presents the results of all the ablation studies. We conduct two branches of ablation studies on ScienceWorld: 

\paragraph{Preplay Analysis.} To demonstrate the effectiveness of our preplay mechanism and soft guidance principle, we first replace preplay with: (1) top-k procedure transition retrieval based on task-specific reliability score (w/ retrieval); (2) random procedure sampling as preplay plan (w/ random sample). Then, we replace the soft guidance with hard constraint, where the agent is forced to strictly follow preplay plan (w/ hard constraint). The results indicate that grounded reasoning or speculative reasoning alone is not always effective. The reasoning flexibility ProPlay provides is necessary. 

\paragraph{World Model Analysis.} To demonstrate the superiority of our procedural world model, we first remove the whole procedural graph (w/o graph), procedure transitions (w/o transition), reliability records (w/o reliability) respectively, and then replace the procedural world model with action-level world model, where actions are regarded as fundamental unit of the world model, while other components except for procedure induction remains the same. The results indicate that procedure transition is the most valuable knowledge within the world model, and our proposed reliability records also contribute to reliable world knowledge exploitation. Besides, procedural world model can provide more comprehensive environment understanding than action-level world model. 

\begin{figure}[!t]
    \centering
    \includegraphics[width=0.9\linewidth]{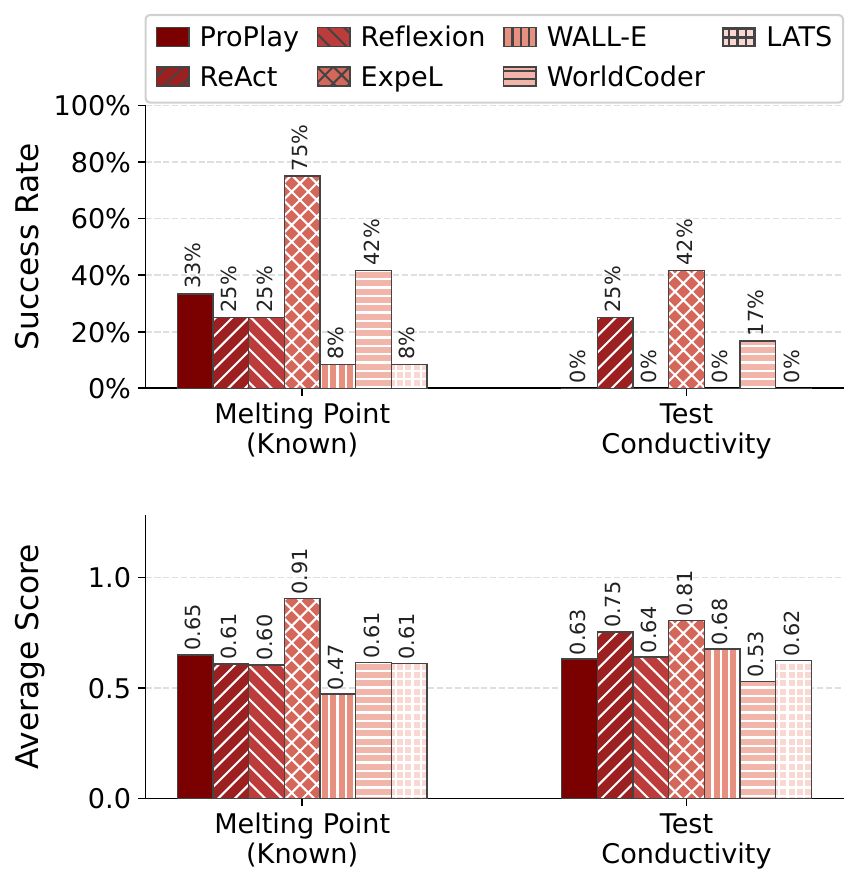}
    \caption{\textbf{Failure Case Analysis on ScienceWorld.} ProPlay underperforms on tasks that require precise quantitative reasoning or non-decomposable tool-assembly sequences.}
    \label{fig:fail}
\end{figure}

\begin{figure}[!t]
    \centering
    \includegraphics[width=\linewidth]{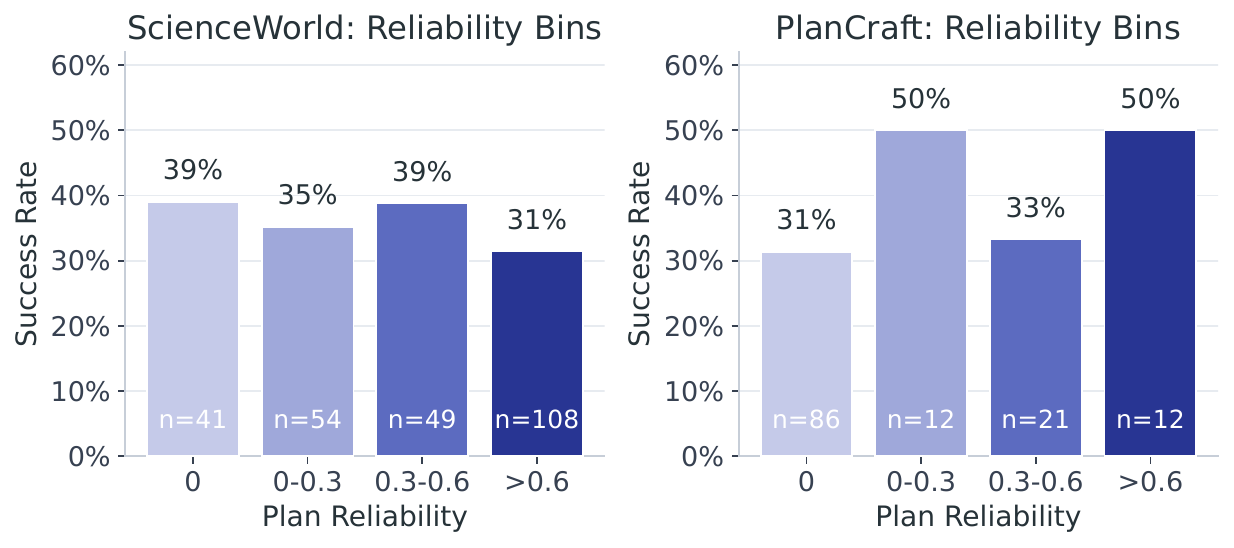}
    \caption{\textbf{Reliability Analysis of ScienceWorld and PlanCraft.} High plan reliability does not necessarily lead to task completion.}
    \label{fig:reliab_sci}
\end{figure}

\begin{figure}[!t]
    \centering
    \includegraphics[width=\linewidth]{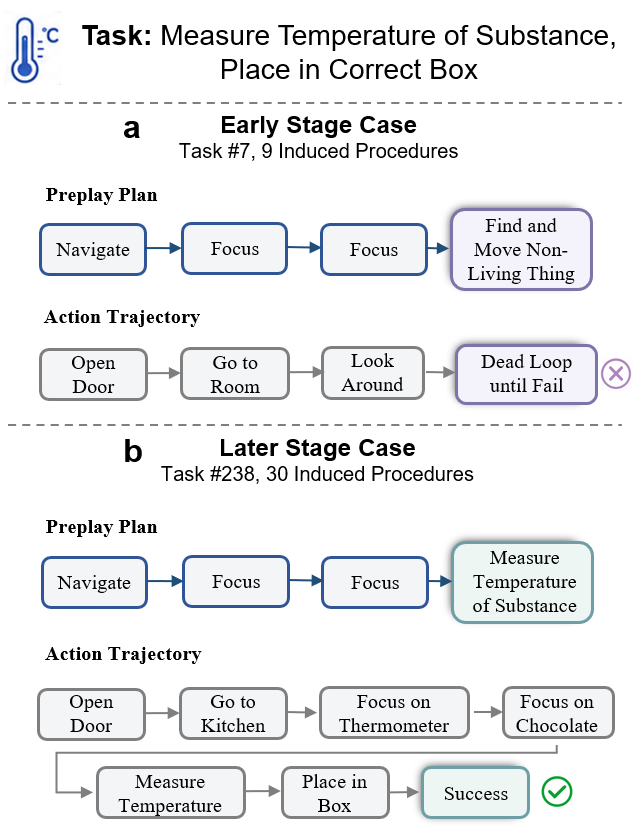}
    \caption{\textbf{Case Analysis on ScienceWorld.} ProPlay can autonomously evolve feasible task-specific procedures.}
    \label{fig:case}
\end{figure}

\subsection{Discussion}
\noindent

\paragraph{Failure Case Analysis.} Figure \ref{fig:fail} visualizes two significant failure cases of ProPlay on ScienceWorld, including melting point measurement (known substance) and conductivity testing. The two cases reveal the fundamental limitations of procedural abstraction. The failure on melting point measurement suggest that tasks requiring precise quantitative reasoning are better served by fine-grained rule-based methods like ExpeL, where procedures may over-generalize across substance-specific thresholds. On the other hand, conductivity testing involves tool assembly, which can not be always decomposed into reusable procedures. A potential solution to overcome the limitation is to integrate complementary fine-grained constraints like action-level rules, enabling the agent to adaptively select the appropriate level of abstraction based on task structure.

\paragraph{Reliability Analysis.} To better understand the role of preplay plans and their reliabilities, we define plan reliability as the average transition reliability score of each procedural plan $\frac{1}{m-1}\sum_{(p_i, p_j)\subset W^*}s_{ij}^k$ and compare plan reliabilities with their corresponding SR. The statistics in ScienceWorld and PlanCraft are summarized in Figure \ref{fig:reliab_sci}. Note that the plans with zero reliability are composed of unseen procedure transitions. 
We can find that high plan reliability does not necessarily lead to high SR and reward. In ScienceWorld, plans with unseen transitions play a critical role, demonstrating the importance of free exploration. In contrast, "more reliable" plans show unexpectedly low SR, which is caused by the existence of high-frequency generic transitions (e.g., navigation $\rightarrow$ focus), which accumulate high global reliability yet offer little task-specific guidance. For PlanCraft, "reliable" plan is also insufficient for task completion, but the contribution of unseen plans is less significant. We attribute such difference to the sparsity of procedure graph, making it difficult to retrieve valuable reference and form high-frequency generic transitions.


\paragraph{Case Study.} We contrast two measure-temperature episodes to demonstrate how the evolution of procedural world model affects plan quality. We select one early episode and one late episode with the same task type in the online inference, and illustrate their preplay plans and actual execution trajectories in Figure \ref{fig:case}. In Task \#7, no temperature-measurement workflow exists, and the plan falls back to a generic "Move Non-Living Thing" procedure with plan reliability 0.14, leading to dead loop and failure. By Task \#238, ProPlay has induced a dedicated "Measure Temperature" procedure with a reliable transition, yielding a feasible task-specific plan with reliability 0.83. As a result, the agent completes the task in 15 steps. 
\section{Conclusion}

We present ProPlay, a preplay framework for self-evolving LLM agents built on an evolving procedural world model. To maintain an internal environment understanding, ProPlay encodes environment dynamics as procedural causal structures, with reliability record embeddings that distinguish generalizable knowledge from contingent experience. Before each episode, ProPlay performs procedure-level preplay to synthesize procedure trajectories as soft guidance, preserving exploration flexibility while grounding agent reasoning in accumulated procedural knowledge. Experiments across diverse benchmarks show that ProPlay consistently outperforms strong baselines, highlighting that procedure-level abstraction offers a principled and effective foundation for self-evolving agent in partially observable environments.

\section*{Limitations}
\noindent

\paragraph{Procedural abstraction scope.} ProPlay's procedural abstraction is most effective for tasks with clear, reusable causal structure. Tasks requiring precise quantitative reasoning and tasks with unpredictable variants are better served by fine-grained rule-based representations. This suggests that procedural structures and action-level rules are complementary, and a principled strategy for adaptively combining them remains an open problem.

\paragraph{Single-Shot Preplay.} The preplay plan is generated once at the start of each episode with only one procedure trajectory, which is injected as static soft guidance throughout execution. When early-episode observations reveal that the plan is incorrect, the agent cannot benefit from a mid-episode plan revision, and must instead rely entirely on its own reasoning to recover. We plan to inject these procedural knowledge into parameters~\citep{han2026survey,han2026w2t} in our future work.

\paragraph{Benchmark Scope.} It remains an open question whether the same benefits of procedural environment modeling can generalize to more open-ended environments like web navigation, GUI interaction, and long-horizon code generation, where the action space is unbounded and procedural structure may not emerge naturally from accumulated experience.

\section*{Ethical Concerns}

This work is entirely verified within publicly available benchmarks (ScienceWorld, $\tau$-Bench, PlanCraft) and does not involve human subjects, private data, or real-world system interactions. The LLM-simulated users in $\tau$-Bench are a proxy for real human behavior and may not reflect the full diversity of actual user responses. We will release all evaluation code, prompt templates and data splits used in our experiments.


\bibliography{custom}

\appendix

\section{Additional Experimental Setting}

We conduct ScienceWorld evaluation based on AgentGym~\cite{xi2025agentgym}, following the corresponding task type settings. The maximum step budget for ScienceWorld, $\tau$-Bench and PlanCraft are set as 100, 30 and 30 respectively. For baseline adaptation, since the original Reflexion operates within a single task with multiple trials, we adapt it to the cross-task setting by accumulating verbal reflections across episodes; at most 20 reflections are retained (FIFO). ExpeL maintains a sliding buffer of up to 50 failed trajectories and extracts at most 20 rules for injection. For LATS, we retain its original within-episode tree search design but limit cross-episode reflection to 3 stored trajectories to bound context length. Wall-E maintains per-verb rule libraries capped at 5 rules each, with a correction buffer of up to 30 examples per verb. WorldCoder and Wall-E were originally designed with offline or multi-trial settings; we adapt both to the online single-trial setting by initializing their world models from scratch at the start of each benchmark run, with no pre-collected demonstrations. For ProPlay, there is no explicit limitation in the length of preplay procedural plan and the size of induced workflow. We set $\delta=0.75$ for both procedure merging and failure experience retrieval. We adopt different procedure definitions for the three benchmarks. For ScienceWorld, we adopt the general procedure definition. For $\tau$-Bench, considering that tools can already represent high-level abstractions, we regard tools themselves as procedures. For PlanCraft, we consider recipes as procedures.

\section{Prompts}

\begin{promptbox}{Procedure Induction}
\textbf{System Message}:

You are maintaining a library of abstract procedure templates for a
task agent. Given the current procedure library and accumulated episode summaries, update the library to reflect generalizable patterns.

Each episode summary contains:

- Actions: the complete action sequence.

- Successful steps: actions from the productive part of the episode.
  Reward-gaining actions are marked with (+delta).
  
- Failed steps (optional): actions attempted after the last reward gain, marked with [x]. Do NOT incorporate these into procedures.

Derive procedure patterns from Successful steps only.

Update rules:

1. ADD a new procedure if episodes demonstrate a task pattern not covered by any existing entry.

2. ADD steps to an existing procedure if episodes reveal steps consistently missing from the general template.

3. REWRITE an existing procedure only if it contains steps that are
   outright incorrect.
   
4. Use abstract placeholders (e.g. <object>, <container>, <location>) — never hard-code values from individual episodes.

5. Do not add conditional branches that apply only to a single episode.

Output:

Section 1 — the complete updated procedure library.

Section 2 — an execution trace for the LATEST episode only: the ordered sequence of procedure names that best describes what the agent actually did. Wrap in <trace> tags, one name per line.

\vspace{1em}
\textbf{User Message:}

\#\# Existing Procedures

\{existing\_procedures\}

\#\# Episodes

\{episode\_summaries\}

\#\# Updated Procedures

\{induced\_procedures\}

\end{promptbox}

\begin{promptbox}{Preplay}

\textbf{System Message:}

You are a planning assistant for an interactive task agent.

Given a task goal and a procedure graph built from past experience,
generate a plan to accomplish the task.

Guidelines:

1. Select only the procedures relevant to this task — not every available procedure needs to be used.

2. Use your own reasoning to determine the order and combination of steps. The procedure graph is evidence, not a prescription.

3. Reliability scores reflect how often a transition contributed to past successful episodes. Higher scores are a useful signal, but not instructions — a low-reliability edge may still be the right choice, and a high-reliability edge may not apply to this task.

4. You may include steps not present in the known procedures if the task requires them.

5. If a Past Episode Experiences section is shown, study each failure entry to identify the root cause, and design your plan to reason around it. Prioritize experiences from tasks similar to the current goal.

Output format:

- Output the plan inside <plan> tags

- Top-level entries are numbered, each beginning with the exact procedure name. Under each entry, list indented sub-steps (-) that adapt the procedure to this specific task.

\vspace{1em}
\textbf{User Message:}

Task Goal

{goal}

Available Procedures

\{procedure\_library\}

Known Procedure Transitions (from past experience)

Sorted by relevance to this task (higher = this sequence worked well on similar tasks).

\{transitions\_with\_reliability\}

Past Episode Experiences

\{failed\_experiences\}

Generate a plan for this task. Use the procedure graph as evidence — follow known sequences where they apply, and add or modify steps based on your understanding of the task.

<plan>
    
\end{promptbox}

\begin{promptbox}{Soft Guidance Injection}

\textbf{System Message:}

Suggested procedure plan for this task (from past experience):
Follow these steps as a guide; adjust based on what you observe.

\{plan\_text\}

\vspace{1em}
\textbf{User Message:}

Task: \{task\}

Observation:
\{current\_observation\}

Working notes (your in-episode memory from previous steps):
\{memory\}

Recent history (last \{k\} steps):
\{history\}

What do you do next?
    
\end{promptbox}

\section{Additional Results}

\subsection{Complementary Task-Specific Results on ScienceWorld}

\begin{figure*}[!t]
    \centering
    \includegraphics[width=\linewidth]{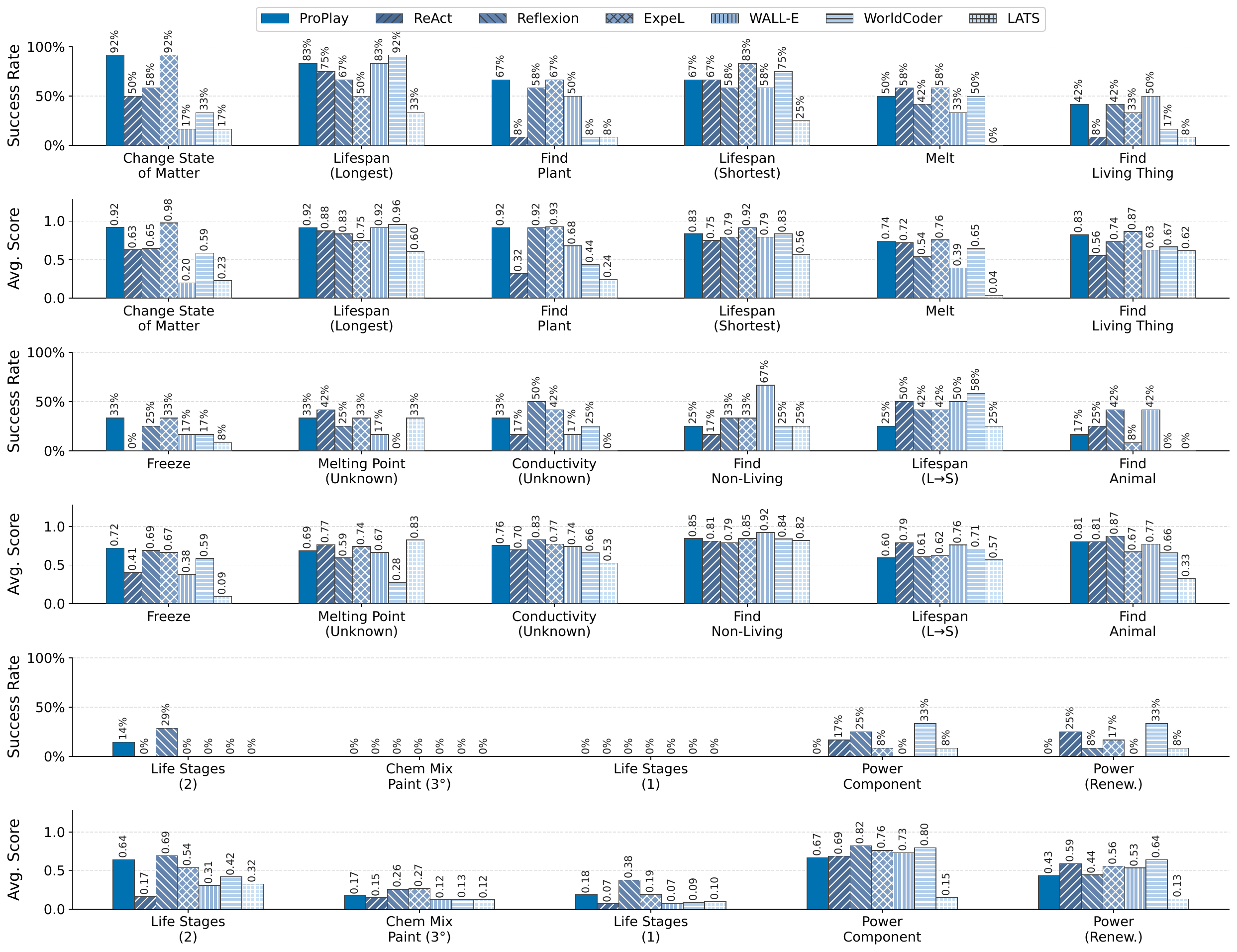}
    \caption{\textbf{Complementary Task-Specific Results in ScienceWorld.}}
    \label{fig:task_all}
\end{figure*}

We summarize the task-specific results of the other 17 tasks in ScienceWorld in Figure \ref{fig:task_all}. The observation is consistent with Figure \ref{fig:task_bar}, such that ProPlay demonstrates consistent advantages on tasks with clear procedural structure and fails on tasks requiring non-decomposable action sequences that are very sensitive to task configurations. 
Specifically, ProPlay matches or leads all baselines on state-transition tasks (change-the-state-of-matter-of, freeze), where the causal chain from action to outcome is fixed and generalizable across episodes. In contrast, ProPlay shows the largest performance gaps on two task families: power-component tasks, where ProPlay achieves 0\% success rate, likely because assembling electrical circuits requires strict action ordering that is highly sensitive to object placement; and open-ended search tasks (find-animal, find-non-living-thing), where the target object location varies unpredictably across configurations, making it difficult for the world model to form reliable relational edges. Compound sequential tasks such as lifespan-longest-then-shortest-lived further expose a limitation: for ranking-based tasks with dependent sub-goals in a strict order, procedural knowledge can not always transfer useful relational knowledge across sub-goals.

\subsection{Procedure Graph Evolution}

\begin{figure*}[!t]
    \centering
    \includegraphics[width=\linewidth]{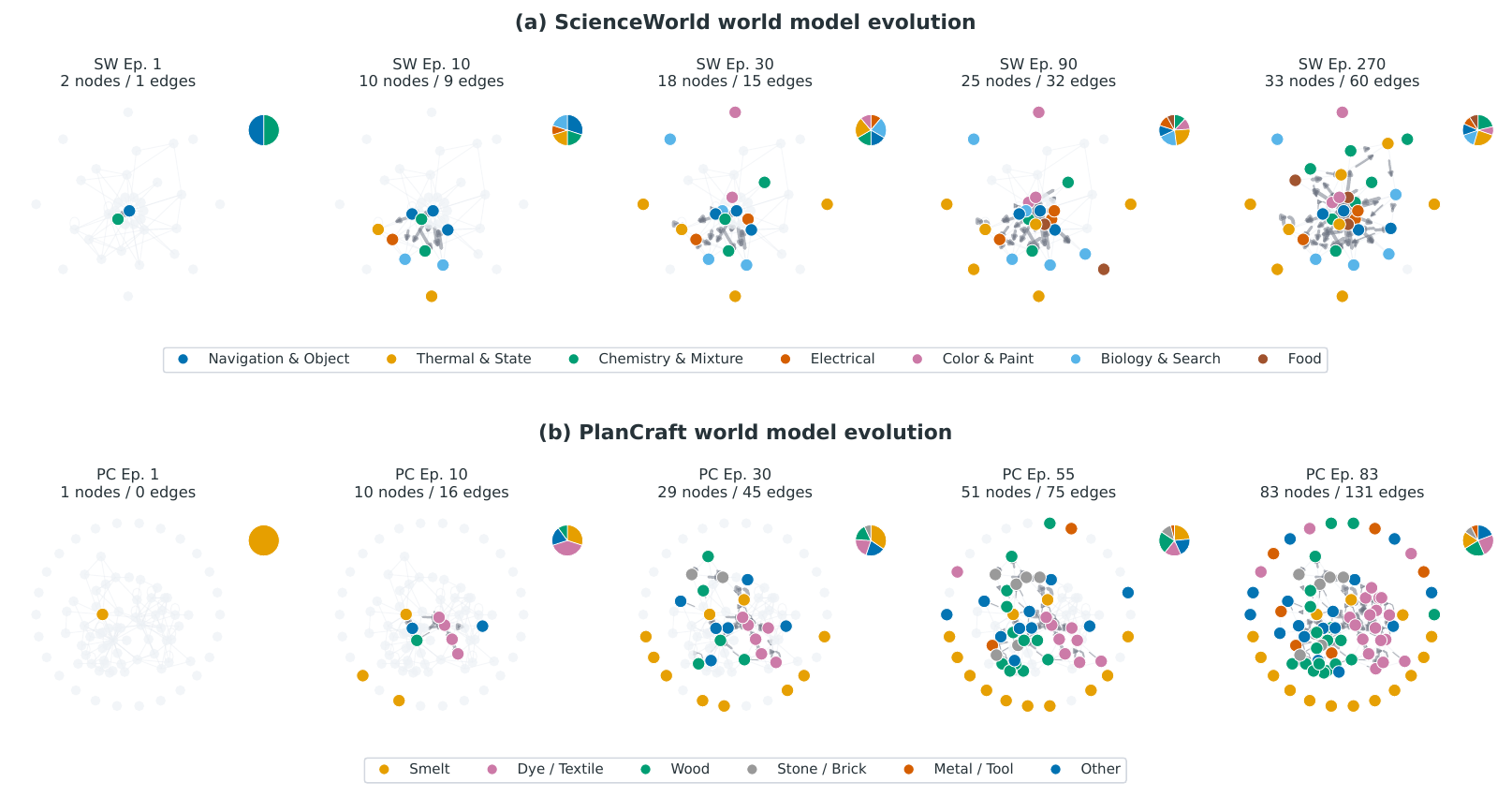}
    \caption{\textbf{Procedure Graph Evolution in ScienceWorld and PlanCraft.} We visualize the procedure graphs induced in episode 1, 10, 30, 90 and 270 respectively, with different node colors representing different targeting tasks.}
    \label{fig:graph_sciworld}
\end{figure*}

We visualize 5 snapshots of evolved procedure graph in ScienceWorld and PlanCraft respectively in Figure \ref{fig:graph_sciworld}. As the graph continues to evolve, the procedural world model in PlanCraft remains a stable edge-to-node ratio (around 1.5) and gradually forms task-specific clusters, reflecting the many-to-one compositional recipe family structure of Minecraft crafting. In contrast, procedures in ScienceWorld are defined as generalizable workflows, where heterogeneous scientific domains all converge on a shared set of manipulation primitives. This structural divergence reflects a fundamental difference in how procedures are defined in each benchmark: PlanCraft procedures are compositionally hierarchical, enabling bottom-up cluster formation, whereas ScienceWorld procedures are induced from common prerequisites, precluding the emergence of domain-isolated subgraphs. This structural contrast suggests that ProPlay can automatically recover the causal structures of different environment dynamics.

\end{document}